\documentclass[letterpaper, 10 pt, conference]{ieeeconf}  

\IEEEoverridecommandlockouts                              

\usepackage{amsmath, amsfonts, amssymb}
\usepackage{bm}
\usepackage{mathtools}
\usepackage{xcolor}
\usepackage{glossaries}
\usepackage{tabularx}
\usepackage{authblk}
\usepackage{algorithm}
\usepackage[noend]{algpseudocode}
\usepackage{tikz}
\usepackage{subcaption}
\usepackage{url}
\usepackage{xfrac}
\usepackage{bbold}
\usepackage{hyperref}
\def\one{\mbox{1\hspace{-4.25pt}\fontsize{12}{14.4}\selectfont\textrm{1}}}
\title{\LARGE \bf
Efficient and Reactive Planning for High Speed Robot Air Hockey
}

\author{Puze Liu$^{1}$,  Davide Tateo$^{1}$, Haitham Bou-Ammar$^{2, 3}$, and Jan Peters$^{1}$\\
\thanks{{$^1$}Department of Computer Science, Technische Universit\"at Darmstadt, Germany \texttt{\{puze,davide\}@robot-learning.de, jan.peters@tu-darmstadt.de}}
\thanks{$^{2}$ Huawei R\&D London, United Kingdom \texttt{haitham.ammar@huawei.com}}%
\thanks{{$^3$} University College London (UCL), Honorary position.}%
}

\newacronym{rl}{RL}{Reinforcement Learning}
\newacronym{drl}{DRL}{Deep Reinforcement Learning}
\newacronym{hrl}{HRL}{Hierarchical Reinforcement Learning}
\newacronym{avi}{AVI}{Approximate Value-Iteration}
\newacronym{api}{API}{Approximate Policy-Iteration}
\newacronym[plural=MDPs, firstplural=Markov Decision Processes (MDPs)]{mdp}{MDP}{Markov Decision Process}
\newacronym{kl}{KL}{Kullback-Leibler Divergence}
\newacronym{gae}{GAE}{Generalized Advantage Estimation}
\newacronym{papi}{PAPI}{Projections for Approximate Policy Iteration}
\newacronym{her}{HER}{Hindsight Experience Replay}
\newacronym{ham}{HAM}{Hierarchy of Abstract Machines}
\newacronym{mom}{MOM}{Measure of Manipulability}
\newacronym{bo}{BO}{Bayesian Optimization}
\newacronym{hebo}{HEBO}{Heteroscedastic Evolutionary Bayesian Optimisation}
\newacronym{ucb}{UCB}{Upper Confidence Bound}
\newacronym{pi}{PI}{Probability of Improvement}
\newacronym{ei}{EI}{Expected Improvement}
\newacronym{nl}{NLP}{Nonlinear Programming}
\newacronym{lp}{LP}{Linear Programming}
\newacronym{qp}{QP}{Quadratic Programming}
\newacronym{aqp}{AQP}{Anchored Quadratic Programming}

\newcommand*{\tran}{^{\mkern-1.5mu\mathsf{T}}}

\def\NullS{{\mathrm{Null}}}
\def\SE{\mathrm{SE}(3)}

\def\vzero{{\bm{0}}}

\def\valpha{{\bm{\alpha}}}

\def\vb{{\bm{b}}}
\def\vc{{\bm{c}}}

\def\vg{{\bm{g}}}

\def\vp{{\bm{p}}}
\def\vq{{\bm{q}}}
\def\vr{{\bm{r}}}

\def\vv{{\bm{v}}}

\def\vx{{\bm{x}}}

\def\mE{{\bm{E}}}

\def\mI{{\bm{I}}}
\def\mJ{{\bm{J}}}

\def\mN{{\bm{N}}}

\def\mW{{\bm{W}}}

\def\NullB{{\mE_\mN}}

\begin{document}

\maketitle
\thispagestyle{empty}
\pagestyle{empty}

\begin{abstract}
Highly dynamic robotic tasks require high-speed and reactive robots. These tasks are particularly challenging due to the physical constraints, hardware limitations, and the high uncertainty of dynamics and sensor measures. To face these issues, it's crucial to design robotics agents that generate precise and fast trajectories and react immediately to environmental changes.
Air hockey is an example of this kind of task. Due to the environment's characteristics, it is possible to formalize the problem and derive clean mathematical solutions. For these reasons, this environment is perfect for pushing to the limit the performance of currently available general-purpose robotic manipulators.
Using two Kuka Iiwa 14, we show how to design a policy for general-purpose robotic manipulators for the air hockey game. We demonstrate that a real robot arm can perform fast-hitting movements and that the two robots can play against each other on a medium-size air hockey table in simulation. 
\end{abstract}

\section{INTRODUCTION}
Recent hardware and software advances allow robots to depart static industrial surroundings to novel real-world \emph{dynamic} environments with (potentially) fast-moving objects interaction involved. In such contexts, real-world constraints, estimation uncertainties, and dynamical interactions pose many challenges. For instance, hardware restrictions limit the design of high-speed trajectories, target tracking errors can lead to complete task failures, and fast-moving objects increase the likelihood of collisions with other obstacles or humans. Therefore, the design of swift, precise, and reactive movements is key to the success of such systems equipping robots with the essential tools to cope with ever-changing conditions. 

Although designing task-specific robotic solution is always a possibility to tackle the aforementioned challenges, empowering \emph{general-purpose} robots to solve dynamic tasks is desirable when the task is not well specified.  Such a robot could face dangerous tasks performed previously by a human if equipped with appropriate end-effectors. However, the progress towards effective general-purpose robotic systems in dynamic environments has been limited due to a variety of factors. A possible step towards better understanding those limitations is to consider a more restricted dynamic environment, such as air hockey.

\begin{figure}[t]
    \centering
    \includegraphics[width=0.48\textwidth, trim=0 0 0 18cm, clip] {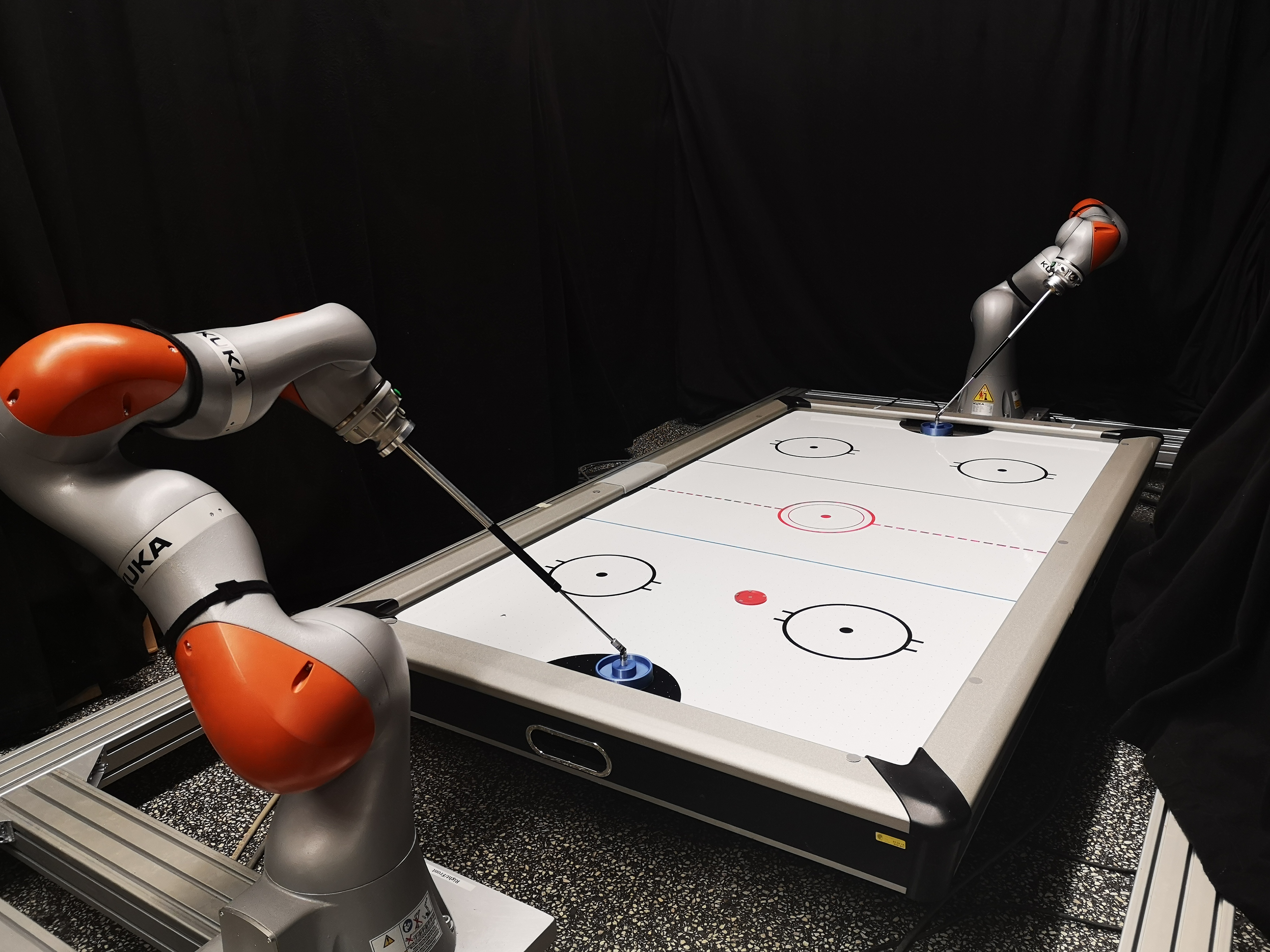}
    \caption{Robot air hockey using two KUKA LBR IIWA 14.}
    \label{fig:air_hockey}
    \vspace{-0.5cm}
\end{figure}

The air hockey task is a 2D constrained environment characterized by fast puck movements and high uncertainty. Two factors cause this high uncertainty: firstly, air flows through small holes on the table surface to reduce the friction between the puck and the table, creating an uneven airflow distribution, resulting in high uncertainty and fast movements. Secondly, the collision behavior between the cylindrical puck and mallet or the table's borders produces highly variable trajectories, as it is sensitive to small differences in the system state. 
This environment shows all the fundamental aspects of dynamic tasks: it requires robots to perform high-speed trajectories that reach the robot capability with low reaction time.
However, it is relatively simple and controlled, letting us formalize the problem rigorously and perform appropriate scientific validation.  

Although many previous works have tried to solve this task~\cite{Partridge2000,Namiki2013,shimada2017two}, few have focused on general-purpose manipulators~\cite{AlAttar2019} and they are not able to show a highly dynamic behavior. Instead, our objective is to show how a general-purpose arm can achieve performances close to the task-specific 2DoF robots by employing advanced optimization techniques. To prove our claim, we consider a real air hockey system with the table size $216 \text{cm} \times 122 \text{cm}$. We mount two KUKA LBR IIWA 14 arms at each end of an air hockey table. The robots play against each other using a two-level policy. The high-level policy selects the appropriate tactic i.e., hitting, defending, etc., while the low-level one computes the required trajectory using planning and optimization techniques.

The main contribution of this work is a novel trajectory optimization technique. We employ a null space optimization algorithm that leverages the robot redundancy to generate high-speed motion while satisfying the joints' position and velocity constraints. We prove the effectiveness of this technique in real robot puck hitting. We also show the two robot arms playing reasonably fast against each other in a simulator. Finally, we use \gls{bo} to tune the physical parameters of the simulation. This approach does not require gradients, enabling us to use standard simulator platforms and easily consider non-differentiable dynamics.

\subsection{Problem Statement}
Air Hockey is a two players game. We refer to the two players as \textit{home} and \textit{away}.
Let $\mathcal{T}\subset\mathbb{R}^2$, be the planar surface of a rectangular Air Hockey Table with dimension $l\times w$, with $w < l$. The table is divided symmetrically in two surfaces, each of length $\frac{l}{2}$, $\mathcal{T}_\text{home}$ and $\mathcal{T}_\text{away}$. Let $\mathcal{G}_\text{home},\mathcal{G}_\text{away}\subset\mathbb{R}^2$ be two players' goal areas, placed at the two short sides of the table.
Let $\vr\in(\mathcal{T}\cup \mathcal{G}_\text{home} \cup \mathcal{G}_\text{away})$ and $\phi\in\left(-\pi, \pi\right]$ be, respectively, the position and orientation of a circular puck of radius $R_\text{mallet}$. Let $x_\text{home},x_\text{away}$ be the position of players' mallets, both of radius $R_\text{mallet}$.
Each player $p$ can modify the puck trajectory, while the puck is in his own area $\mathcal{T}_\text{p}$, by causing a collision between the mallet and the puck i.e., $\lVert \vr - \vx_{p} \rVert_2 = R_\text{puck} + R_\text{mallet}$. The puck can change direction also by colliding with the boundaries of the table, excluding the boundaries between $\mathcal{T}$ and the goal areas $\mathcal{G}$.
The objective of each player $p$ is to maximize their score 
\begin{equation}
\max \sum_{i}^{N} \sum_t^{T_i}  \mathbb{1}(\vr_t\in\mathcal{G}_{\lnot p}),
\nonumber
\end{equation}
where $N$ is the number of game rounds, $T_i$ is the length of round $i$, and $\lnot p$ is the opponent.

Given that the mallet's position should stay on the table surface to hit effectively, a robotic agent must be able to execute fast trajectories on the planar manifold. As this task is tremendously difficult at high speed, the objective is to minimize the trajectory error w.r.t. the plane surface in the $z$ axis. We add a mechanical compliant end-effector to cope with the error.

\subsection{Related Work}
Robotics researchers have a long-standing interest in Air Hockey. Such interest lies in the fact that Air Hockey is a very dynamic game that can exhibit interesting robot dynamics and tactics. 

One of the first examples of a robotic Air Hockey system is presented in~\cite{bishop1999vision}, where the authors designed a planar, 3-link, redundant manipulator that can play using cameras. In a subsequent paper~\cite{Partridge2000}, the authors described in detail the dynamics of the game, focusing specifically on collision modeling.
One of the most outstanding Air Hockey platforms has been presented in~\cite{Namiki2013}, where a 4Dof Barrett Wam arm has been used as a planar robot to play air hockey fast. In the paper, the authors present a hierarchical architecture that considers low-level control as well as short and long-term strategies. 
It is possible to find another example of a planar robot playing Air Hockey  in~\cite{shimada2017two}, where the authors implement a two-layer tactical system.
The only known general-purpose manipulator to solve this task can be found in~\cite{AlAttar2019}, where the authors use a Franka Emika Panda Arm. The proposed tactical system is strongly inspired on~\cite{Namiki2013} and~\cite{shimada2017two}. However, the system's performance is quite far from the task-specific ones, which use robots in a planar configuration.
Up to our knowledge, our platform is the only general-purpose manipulator able to play air hockey with reasonable performances on a medium-size table.

Many works focus on simulated Air Hockey environments. In~\cite{Alizadeh2013}, the authors developed an automatic calibration procedure that allows the system to automatically tune the parameters that affect the puck's motion e. g., the restitution coefficients.
In~\cite{Igeta2017}, the authors have studied how to optimize the attack motion against a human opponent. In~\cite{Taitler2017}, the authors developed a modified version of the DQN algorithm~\cite{mnih2015human}, incorporating suboptimal demonstrations. The authors show that the proposed approach outperforms other deep learning baselines.

The air hockey task has also been used in the Robot Learning community. An example is~\cite{Xie2020}, where the authors use a very simplified air hockey task as a benchmark. In this work, the authors learn a latent representation of the opponent's tactics. However, the used environment is extensively limited and not comparable to other works that focus on the air hockey task. These limitations highlight the need for a fully functional air hockey platform. Thus, a more complex environment is also beneficial for the Robot Learning community. This task can be used as a test bench for many different Robot Learning areas: ranging from learning dynamical movements to multiagent learning and human-robot interaction.

\section{PRELIMINARIES}

\subsection{Manipulator kinematics} 
We consider general-purpose, redundant, 7DoF manipulators. A robot configuration is fully described by the joint angles. The joint space is the space of all the possible joint angles:
$\vq=\lbrace q_i | i\in\lbrace1\dots7\rbrace,\, q_{i}^{\min} \leq q_i \leq  q_{i}^{\max}  \rbrace.$
We can compute the end-effector pose $\vx$ in a given configuration $q$ by using the forward kinematics i.e., $\vx=\text{FK}(\vq)$. The space of all allowed end-effector poses is called the task space. For any pose in the task space, we can compute one (or multiple) configurations in terms of joint positions using the inverse kinematics i.e., $\vq=\text{IK}(\vx)$. For a redundant 7DoF manipulator, $\vq\in\mathbb{R}^7$ and $\vx \in \SE$. In this setting, the end-effector pose does not fully describe the system's configuration due to the extra degree of freedom originating from the robot's redundancy.
From this fact, it follows that, for most configurations, the end-effector pose does not fully describe the system's configuration. There is still a degree of freedom due to the robot's redundancy.

The relation between the joint velocities $\dot{\vq}$ and task space velocities $\dot{\vx}$ is given through the Jacobian Matrix $\dot{\vx} = \mJ(\vq) \dot{\vq}. \label{eq:diff_ee_vel}$ For a redundant manipulator, it is possible to compute the desired joint velocities given a target end-effector velocity by solving \eqref{eq:diff_ee_vel} as follows
\begin{equation}
    \dot{\vq} = \mJ^{\#}(\vq) \dot{\vx} + \mN \dot{\vq}, \label{eq:diff_joint_vel}
\end{equation}
with the generalized inverse $\mJ^{\#}$ and the null space projection matrix $\mN = \mI - \mJ^{\#}\mJ$. A common choice of the generalized inverse is the pseudoinverse $\mJ^\dagger = \mJ\tran (\mJ \mJ\tran)^{-1}$. Here, the second component describes the joint velocities that change the robot's redundancy while fixing the end-effector pose.

Instead of using the null space projecting matrix, it is possible to parameterize the null space velocities using the coordinate system induced by the null space basis. We can rewrite \eqref{eq:diff_joint_vel} as
$\dot{\vq} = \mJ^\#(\vq) \dot{\vx} + \NullB \valpha, $
where $\valpha$ is the velocity in the null space coordinates, $\NullB=\NullS(\mJ)$ is the matrix of basis vectors spanning the null space of $\mJ$. When considering the 7DoF robot and the end-effector $\vx\in\SE$, the null space matrix becomes a single basis vector, $\valpha$ becomes a scalar. This formulation is particularly useful to perform non-linear optimization.

An important metric that can be used to choose a given configuration is the \gls{mom} $w=\sqrt{\lvert \mJ(\vq) \mJ\tran(\vq) \rvert}$.
The \gls{mom} metric describes the distance to the singularity, as it is non-negative and becomes zero only at the singularities. \gls{mom} is also useful to understand in which configuration the joint velocities produce the highest end-effector velocity.

\subsection{Bayesian Optimization} \label{Sec:BO}
We use \gls{bo} as a tool for portable and flexible system identification. In \gls{bo}, the objective is to \emph{efficiently} maximize a (stochastic) black-box model based on function-value information, i.e., $\max_{\bm{\theta} \in {\Theta}} f(\bm{\theta})$ with $\Theta \subseteq \mathbb{R}^{d}$ being the $d$-dimensional search domain. Algorithms of this type are sequential in nature. At each round $i$, an input $\bm{\theta}_{i} \in \Theta$ is selected and a black-box function value $f(\bm{\theta}_{i})$ is observed. The goal is to rapidly (in terms of regret) approach $\bm{\theta}^{\star} = \arg\max_{\bm{\theta} \in \Theta} f(\bm{\theta})$. Since both $f(\cdot)$ and $\bm{\theta}^{\star}$
are unknown, solvers need to trade off exploitation and exploration during the search process. 

To achieve this goal, typical \gls{bo} algorithms operate in two steps. In
the first, a Bayesian model is learned, while in the second
an acquisition function determining new queries is
maximized. Next, we briefly survey both those steps. 
\subsubsection{BO with Gaussian Processes (GPs)}\label{Sec:GP} Gaussian process regression~\cite{Carl} offers a flexible and sample efficient modelling alternative to reason about $f(\cdot)$. In those forms, designers impose a GP prior on latent functions, which are fully specified by a mean function $m(\bm{\theta})$ and a covariance kernel $k_{\bm{\lambda}}(\bm{\theta}, \bm{\theta}^{\prime})$ with $\bm{\lambda}$ being kernel hyper-parameters. Assuming Gaussian likelihood noise with variance $\sigma$ and given a data-set $\mathcal{D}_{i} = \{\bm{\theta}_{l}, y_{l}\equiv f(\bm{\theta}_{l})\}_{l=1}^{n_{i}}$ with $n_i$ denoting all gathered data up to the $i^{th}$ round, one can compute output predictions on novel input queries $\bm{\theta}_{1:q}^{\star}$. This is achieved through the predictive posterior which is given by $f(\bm{\theta}^{\star}_{1:q})|\mathcal{D}_{i}, \bm{\lambda} \sim \mathcal{N}\left(\bm{\mu}_{i}(\bm{\theta}_{1:q}^{\star};\bm{\lambda}), \bm{\Sigma}_{i}(\bm{\theta}_{1:q}^{\star};\bm{\lambda})\right)$ with
\begin{align}
\label{Eq:muandSig}
    \bm{\mu}_{i}(\bm{\theta}_{1:q}^{\star};\bm{\lambda}) & = \overbrace{\bm{K}^{(i)}_{\bm{\lambda}}(\bm{\theta}_{1:q}^{\star}, \bm{\theta}_{1:n_{i}})}^{\bm{A}^{(i)}}\tilde{\bm{K}}_{\bm{\lambda}}^{(i)}\bm{y}_{1:n_{i}},
    \\ \nonumber
    \bm{\Sigma}_{i}(\bm{\theta}_{1:q}^{\star};\bm{\lambda})&= \bm{K}^{(i)}_{\bm{\lambda}}(\bm{\theta}_{1:q}^{\star}, \bm{\theta}_{1:q}^{\star}) -  \bm{A}^{(i)}\tilde{\bm{K}}^{(i)}_{\bm{\lambda}} \bm{A}^{\mathsf{T}, (i)},
\end{align}
where we have used $\bm{\theta}_{1:n_{i}}$ and $\bm{\theta}_{1:q}^{\star}$ to concatenate all $n_{i}$ inputs and $q$ queries, and $\tilde{\bm{K}}_{\bm{\lambda}}^{(i)} = [\bm{K}_{\bm{\lambda}}(\bm{\theta}_{1:n_{i}}, \bm{\theta}_{1:n_{i}}) + \sigma \bm{I}]^{-1}$. In the latter $\bm{K}_{\bm{\lambda}}(\bm{\theta}_{1:n_{i}}, \bm{\theta}_{1:n_{i}}) \in \mathbb{R}^{n_{i} \times n_{i}}$ denotes the covariance matrix that is computed such that $[\bm{K}_{\bm{\lambda}}(\bm{\theta}_{1:n_{i}}, \bm{\theta}_{1:n_{i}})]_{k,l} = k_{\bm{\lambda}}(\bm{\theta}_{k}, \bm{\theta}_{l})$. 

The remaining ingredient needed in a GP pipeline is a process to determine the unknown hyper-parameters $\bm{\lambda}$ given a set of observations. In standard GPs~\cite{Carl}, $\bm{\lambda}$ are fit by minimising the negative log likelihood leading us to the following optimization problem
\begin{align*}
    \min_{\bm{\lambda}} \mathcal{J}(\bm{\lambda}) = \frac{1}{2} \text{det}\left(\tilde{\bm{K}}_{\bm{\lambda}}^{(i)}\right) + \frac{1}{2}\bm{y}^{\mathsf{T}}_{1:n_{i}}\tilde{\bm{K}}_{\bm{\lambda}}^{(i)}\bm{y}_{1:n_{i}} + \frac{n_{i}}{2} \log 2\pi. 
\end{align*}
\subsubsection{Acquisition Function Maximization}\label{Sec:Acq} Acquisition functions trade off exploration and exploitation for determining new probes to evaluate by utilising statistics from the posterior $p_{\bm{\lambda}}(\cdot) \equiv p(f(\cdot)|\mathcal{D}_i, \bm{\lambda})$. Among various types, we focus on three myopic acquisitions being \gls{ei}~\cite{movckus1975bayesian}, \gls{pi}~\cite{kushner1964new}, and \gls{ucb}~\cite{srinivas2009gaussian}. \\
\underline{\textbf{EI Acquisition:}} In \gls{ei}, one determines new queries by maximizing expected gain relative to the function values observed so far~\cite{movckus1975bayesian}. In a batch form, an \gls{ei} acquisition is defined as
\begin{equation*}
    \alpha_{\text{EI}}(\bm{\theta}_{1:q}|\mathcal{D}_i) = \mathbb{E}_{p_{\bm{\lambda}}(\cdot)}\left[\max_{j\in1:q}\left\{\text{ReLU}(f(\bm{\theta}_{j})-f(\bm{\theta}_{i}^{+}))\right\}\right],
\end{equation*}
where $\bm{\theta}_{j}$ is the $j^{th}$ vector in the batch $\bm{\theta}_{1:q}$, $\bm{x}^{+}_i$ is the best performing input so far and $\text{ReLU}(a) = \max\{0, a\}$.  
\underline{\textbf{PI Acquisition:}} To assess a new batch of probes, \gls{pi} measures the probability of acquiring gains in the function value compared to $f(\bm{\theta}_{i}^{+})$. Such a probability is measure through an expected left-continuous Heaviside function, $\one (\cdot)$, as follows
\begin{equation*}
    \alpha_{\text{PI}}(\bm{\theta}_{1:q}|\mathcal{D}_i) = \mathbb{E}_{p_{\bm{\lambda}}(\cdot)}\left[\max_{j\in1:q}\left\{\one\{f(\bm{\theta}_j) - f(\bm{\theta}^{+}_{i})\}\right\}\right],
\end{equation*}
\underline{\textbf{UCB Acquisition:}} In this type of acquisition, the learner trades off the mean
and variance of the predictive distribution to gather new query points for function evaluation
{\small 
\begin{equation*}
     \alpha_{\text{UCB}}(\bm{\theta}_{1:q}, \mathcal{D}_i) = \mathbb{E}_{p_{\bm{\lambda}}} \left[\max_{j\in1:q}\{\mu_{i}(\bm{\theta}_{j}; \bm{\lambda}) + \sqrt{\sfrac{\beta \pi}{2}}|\gamma_{i}(\bm{\theta}_{j};\bm{\lambda})|\}\right],
\end{equation*}
}
where $\mu_{i}(\bm{\theta}_{j}; \bm{\lambda})$ is the posterior mean given in Equation~\ref{Eq:muandSig} and $\gamma_{i}(\bm{\theta}_{j};\bm{\lambda}) =f(\bm{\theta}_{j}) - \mu_{i}(\bm{\theta}_{j}; \bm{\lambda})$.

With modeling and acquisition ingredients defined, \gls{bo} runs a loop that iteratively fits a GP model (Section~\ref{Sec:GP}) and then determines a batch of new probes to evaluate by maximizing acquisitions from Section~\ref{Sec:Acq}.   

\section{OPTIMIZATION OF HITTING MOVEMENT}\label{Sec:Hit}
Hitting is the most challenging movement to perform in the air hockey task.
It is necessary to accelerate the puck to high velocities, such that two robots can play autonomously against each other.
To obtain this result, we need to maximize the mallet's speed when the collision with the puck happens.

Unfortunately, it can be hard to achieve this objective with a general-purpose, 7-DoF manipulator.
While the maximum joint velocity can be sufficient for high-speed point-to-point movements, the environment's constraints, i.e., the plane manifold of the air hockey game, introduce many problematic issues. Specifically, these constraints are located in two different spaces (the task space and the joint space). The classical approach consists of planning and inverse kinematics. This method solves the problem in the task space and doesn't consider the joint velocity constraints. Thus, it is not suitable for high-speed movements. Conversely, optimizing in the joint space leads to many high-dimensional constrained non-linear optimization problems, which do not fit the real-time requirements.

In our approach, we plan a collision-free Cartesian trajectory based on the start, hitting, and stop points such that the Cartesian constraints are satisfied. Then, we propose a linear constrained \gls{qp} to compute desired joint velocities on each trajectory point. Unlike the simple joint trajectory optimization method, where both objective and constraints are nonlinear, we can compute the solution of \gls{qp} fast and easily. With our approach, we can find a hitting trajectory in real-time satisfying both Cartesian and joint constraints. 

Moreover, we optimize the joint configuration at the hitting point to maximize the robot's performance.
At the hitting point, instead of constructing a non-linear, high-dimensional (position + velocity) optimization problem, we decompose the problem into two lower-dimensional problems. Firstly, we perform a position-only \gls{nl} by maximizing the manipulability along the hitting direction. Secondly, we propose two methods to find the maximum end-effector velocity reachable in that configuration. Given the desired hitting configuration, we present the \gls{aqp}, a variant of the original \gls{qp}, which significantly improves the hitting performance.

\subsection{Cartesian Trajectory Planning}
\begin{figure}[t]
    \centering
    \includegraphics[width=\columnwidth, trim=0cm 2cm 0cm 2cm, clip]{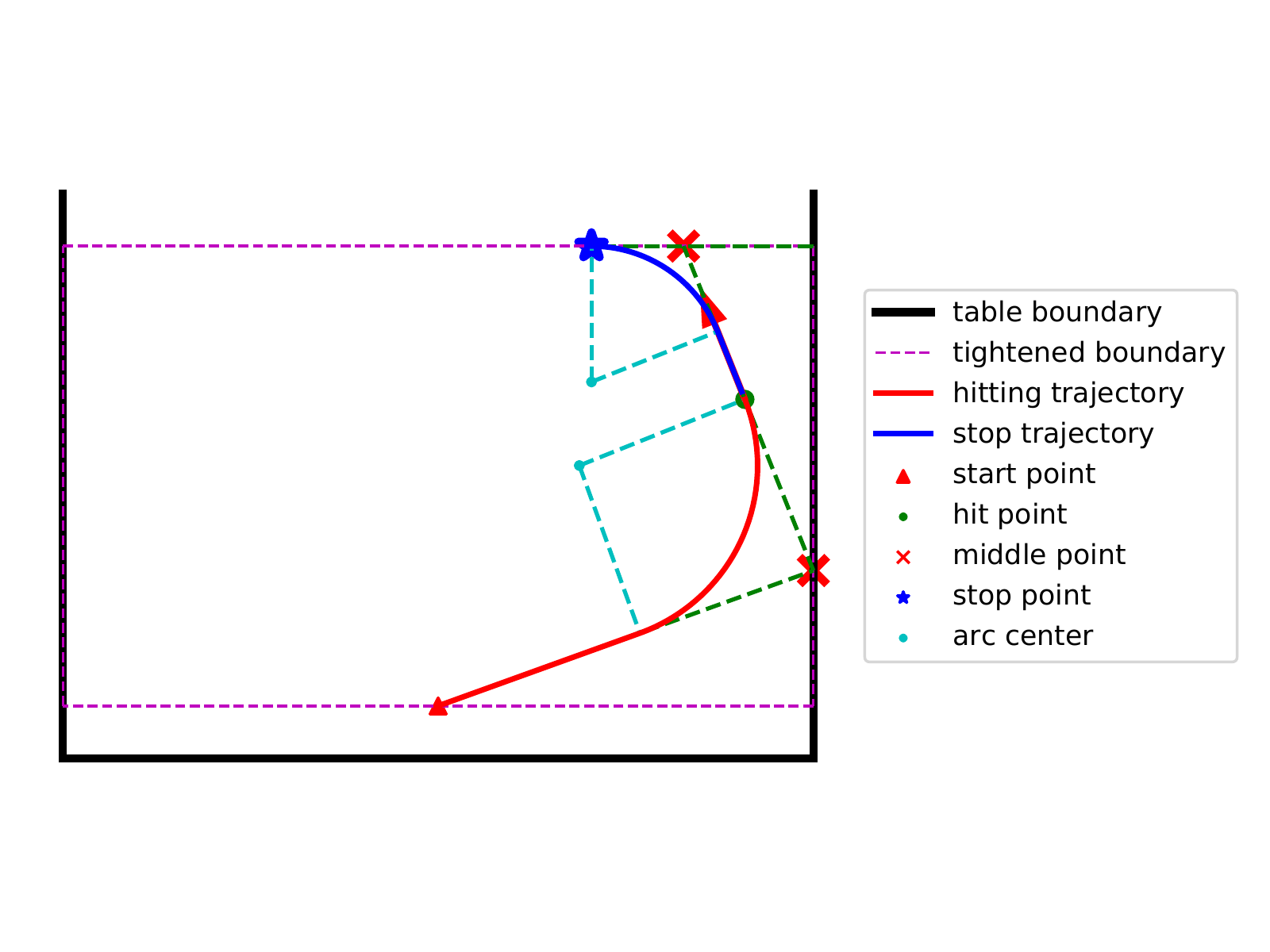}
    \caption{Collision-free hitting path planning. We use two auxiliary middle points to find the path. The arc radius is maximized to reduce angular acceleration.}
    \label{fig:movement_planning}
\end{figure}

Typical trajectory planning methods are not applicable in the air-hockey task, for example, the cubic/quintic polynomials could exceed the table's boundaries, the Bezier curve \cite{biagiotti2008trajectory} does not provide a monotonic velocity profile (the maximum velocity doesn't occur at the hitting point). Therefore, we design a trajectory planning method composed of two segments: \textit{hitting}, \textit{stop}. As illustrated in Fig. \ref{fig:movement_planning}, each segments of the trajectory is composed of two parts: \textit{linear part} and \textit{arc part}. Given the start and the stop point, the tightened boundaries are firstly determined to prevent a motion in an unnecessary direction. Then we find two cross-points between the tightened boundaries and the line that passes through the hit point along the hitting direction. The \textit{arc part} is tangent to the line segment of start/endpoint and middle point as well as the line segment of middle point and hit point with the maximum arc radius. The \textit{linear part} links the arc to the unconnected point.

The planed path is parameterized by the arc length $\vx = \vg(s)$. For each segment, we use a quartic polynomial to define the profile of the arc length $s(t) = a_0 + a_1 t + a_2 t^2 + a_3 t+3 + a_4 t^4$. We employ the positions and velocities boundary conditions as the desired position and velocity of each segment and set the acceleration boundary conditions at two endpoints as $\ddot{s}(0) = \ddot{s}(t_f) = 0$, the arc length is thus guaranteed to be monotonic in the time interval $[0, t_f]$.The total motion time is $t_f = 2 s_f / v_{\text{hit}}$, where $v_{\text{hit}}$ is the speed at one of the two endpoints that is greater than 0.

\subsection{Weighted Null Space Optimization (\texorpdfstring{\gls{qp}}{QP})}\label{sec:qp}
To achieve high-speed trajectories, we optimize the null space velocity at every point of the trajectory. We attach a universal joint to the end-effector (Section~\ref{Sec:Hardware}) and focus purely on the positions of the end-effector in task space, the concerned Jacobian reduced to $\mJ_p \in \mathbb{R}^{3 \times n}$.
Let $\vb = \mJ_{\vp}^{\dagger} \left(\dfrac{\bar{\vx} - \vx}{\Delta T}  + \bar{\vv}\right)$ be the joint velocities for the trajectory tracking, $\vx, \bar{\vx}, \bar{\vv}$ are respectively actual, desired cartesian position and desired cartesian velocity. The objective is to minimize the weighted joint velocity as
\begin{align}
  \min_{\bm{\alpha}} \quad & \dfrac{1}{2} \left\Vert \vb + \NullB\bm{\alpha} \right\Vert^2_\mW \nonumber \\
  \text{s.t.} \quad & \dot{\vq}^{\min}\leq \vb + \NullB \bm{\alpha} \leq \dot{\vq}^{\max} \label{eq:null_space_opt},
\end{align}
where $\mE_{\mN} \in \mathbb{R}^{n\times (n-3)}$ is the \textit{null space matrix} that each column is one basis vector of the $\mathrm{Null}(\mJ_{\vp})$ and $\bm{\alpha} \in \mathbb{R}^{n-3}$ are entries on each basis vector. 

It should be pointed out that, when the $\mW$ is set to be an identity matrix, the optimal solution is $\bm{\alpha} =\vzero$ when the constraints are not violated. In practice, we apply higher weights on the shoulder and the elbow joints because these joints afford more loads and have lower velocity limits.
Finally, we can obtain the next step velocity and position by Euler integration
\begin{align*}
    \dot{\vq}_\text{next} & = \vb + \NullB \bm{\alpha}^* , \nonumber\\
    \vq_\text{next} & = \vq_\text{current} + \dot{\vq}_\text{next}\Delta t.
\end{align*}

\subsection{Hitting Configuration Optimization (\texorpdfstring{\gls{nl}}{NL})}
To find the best configuration for the hitting movement, we consider the manipulability along a specific hitting direction $\vv$, as presented in~\cite{vahrenkamp2012manipulability}.
We maximize the following quantity at the hitting point $\vp$
\begin{align}
    \max_{\vq}  \; \left\Vert (\vv\tran \mJ_p (\vq))\right\Vert_2, \qquad
    \mathrm{s.t.} \; \mathrm{FK}_p(\vq) = \vp, \label{eq:nl_opt} 
\end{align}
And $\mathrm{FK}_p (\vq)$ is the forward kinematics w.r.t positions $(x, y, z)$.

\subsection{Computing the Maximum Hitting Velocity}\label{sec:VelOpt}
To compute the maximum hitting velocity at the selected hitting configuration, we can use two different strategies: 1) the least square solution, which can be computed easily in closed form 2) the linear programming approach. The least-square approach produces a feasible solution while the linear programming method results in the maximum theoretical velocity on that configuration.

\paragraph{Least square solution} We parameterize the hitting velocity as a scalar speed value $\eta$ and a hitting direction $\vv$, such that $\vv_{\max} = \eta \vv$.
The least square solution is
$\dot{\vq} = \mJ_\vp^\dagger (\vq ^*) \vv_{\max}. $
Thus, the maximum possible joint velocity can be determined by the minimum ratio of the absolute value of $i$-th joint velocity $\dot{q}_i$ and its maximum $\dot{q}_{i}^{\max}$
\begin{equation*}
    \eta = \min_i\left(\frac{\dot{q}_{i}^{\max}}{|\dot{q}_i|}\right), \quad i \in \{1, \cdots, n\}.
\end{equation*}

\paragraph{\gls{lp}} Instead of using least squared solution, we can construct a linear program on the null space as 
\begin{align}
    \max_{\valpha} \quad & \vv\tran \mJ_{\vp} (\vq^*) \left(\mE_{\vv^{\perp}}(\vq^*) \valpha \right), \nonumber\\
    \mathrm{s.t.} \quad & \dot{\vq}^{\min} \leq \mE_{\vv^{\perp}}(\vq^*) \valpha \leq \dot{\vq}^{\max}, \label{eq:linear_prog}
\end{align}
where $\vv^\perp \in \mathbb{R}^3$ is the orthogonal complement space of $\vv \in \mathbb{R}^3$, which in our case is a plane orthogonal to the hitting direction, $\mE_{\vv^{\perp}}(\vq^*) = \NullS\left[(\vv ^\perp)\tran \mJ_{\vp} (\vq^*)\right]$ is the basis vectors spanning the null space of $(\vv ^\perp)\tran \mJ_{\vp} (\vq^*)$.

\subsection{Anchored Null Space Optimization (\texorpdfstring{\gls{aqp}}{AQP})}
The \gls{qp} optimization mentioned in Section~\ref{sec:qp} is a local optimizer that focuses only on the current step. It can lead to the local optima with undesirable redundancy configuration, e.g., lower elbow configurations can cause collisions with the table. To avoid this problem, we can modify the optimization problem to an \gls{aqp}
\begin{align}
  \min_{\bm{\alpha}} \quad & \dfrac{1}{2} \left\Vert  \left(\vb + \NullB\bm{\alpha}\right) - \dot{\vq}^a \right\Vert^2_\mW, \nonumber \\
  \text{s.t.} \quad & \dot{\vq}^{\min}\leq \vb + \NullB \bm{\alpha} \leq \dot{\vq}^{\max}, \label{eq:anchor_opt}
\end{align}
where $\dot{\vq}^a = z \frac{(\vq^a - \vq)}{c}$ is a reference velocity that leads to the anchored configuration $\vq^a$ which is solved from hitting configuration optimization and $z$ is the phase variable, which linearly increasing from 0 in the hitting segment and linearly decreasing to 0 in the stop segment and $c$ is a scale constant. Instead of minimizing the weighted joint velocity in~\eqref{eq:null_space_opt}, we minimize the distance to the reference velocity. When $z=0$ at the hitting point, the objective is the same as~\eqref{eq:null_space_opt} and when $z=1$ at the two ends of the trajectory, the objective tries to find the closest configuration to the reference.

\section{SYSTEM IDENTIFICATION} 
\begin{figure}[t]
  \centering
  \includegraphics[width=\columnwidth] {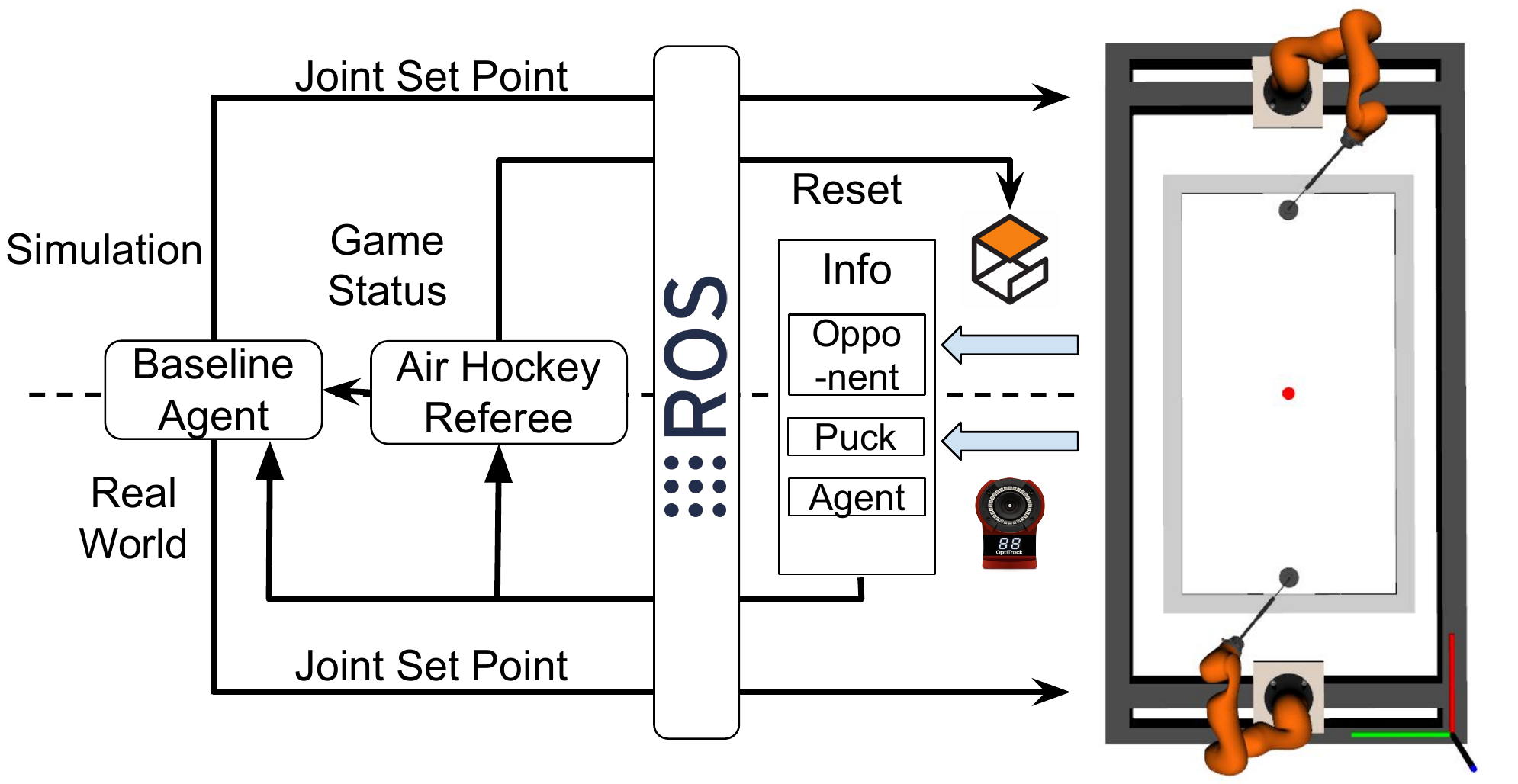}
  \caption{Framework of the air hockey system. Real world and simulation share the same structure.}
  \label{fig:framework}
\end{figure}

\begin{table}[b]
    \centering
    \begin{tabular}{|c|c|c|}
    \hline
    \textbf{parameter} & \textbf{min} & \textbf{max} \\
    \hline
    restitution long-side rim & 0.5 & 1\\
    \hline
    restitution short-side rim & 0.5 & 1\\
    \hline
    friction coefficient rim & 0 & 0.5\\
    \hline
    friction coefficient table-surface & 0 & 0.5\\
    \hline
    puck spinning & 0 & 0.5\\
    \hline
    puck linear velocity decay & 0 & 0.01\\
    \hline
    puck angular velocity decay & 0 & 0.01\\
    \hline
\end{tabular}
    \caption{Parameters for system identification}
    \label{tab:param_system_identification}
\end{table}
For in-depth safe experimentation, we resort to building an accurate simulator of the puck's behavior in Gazebo. We realize a set of 7 tunable parameters that are summarized in Table~\ref{tab:param_system_identification}. Due to the difficulty associated with differentiating through a Gazebo simulator, we follow a Bayesian optimization scheme (Section~\ref{Sec:BO}) for allocating optimal configurations of those parameters. Our black-box function consists of two terms responsible for errors between the puck's position and orientation as generated by Gazebo versus real data gathered using the Optitrack system, i.e., $\{\langle \bm{r}^{(i)}_{t}, \phi^{(i)}_{t}\rangle_{t\geq 1}\}_{i\in [1:N]}$ with $N$ being the total number of collected trajectories. In short, we define the black box as an average over real-world traces as follows:  
\begin{align*}
    {f}(\bm{\theta}) =\mathbb{E}_{i\sim\mathcal{U}[1:N]}\Bigg[\frac{1}{T}\sum_{t=1}^{T}||\Delta r_{t}^{(i)}(\bm{\theta})||_{2}^{2} + 0.2\lVert \Delta \phi^{(i)}_t(\bm{\theta}) \rVert^2_\angle\Bigg],
\end{align*}
where $\Delta r_{t}^{(i)} = \bm{r}^{(i)}_{t} - \hat{\bm{r}}^{(i)}_{t}(\bm{\theta})$ and $\Delta\phi_{t}^{(i)} = \phi_{t}^{(i)} - \hat{\phi}_{t}^{(i)}(\bm{\theta})$ with $\{\langle \hat{\bm{r}}^{(i)}_{t}(\bm{\theta}), \hat{\phi}^{(i)}_{t}(\bm{\theta})\rangle_{t\geq 1}\}_{i\in [1:N]}$ being positions and orientations gathered from Gazebo under a specific setting of the parameters $\bm{\theta}$ from Table~\ref{tab:param_system_identification}. In the above equation we use $\mathcal{U}(\cdot)$ to denote a uniform distribution over the trajectories, while $\angle$ represents the (shortest) angular distance between $\phi^{(i)}_{t}$ and $\hat{\phi}_{t}^{(i)}(\bm{\theta})$.  

Unfortunately, \gls{bo} as described in Section~\ref{Sec:BO} faces two significant challenges when attempting to optimize for $\bm{\theta}$. First, nonlinearities and puck collisions induce a non-Gaussian (heteroscedastic) likelihood noise violating Gaussianity assumptions inherent to standard GPs, and second, the choice of the acquisition to optimize is vague, whereby different acquisitions can lead to conflicting minima~\cite{cowenempirical}. To circumvent the aforesaid problems, we adapt techniques from \gls{hebo}~\cite{cowenempirical} to tune the parameters in Table~\ref{tab:param_system_identification}. Precisely, we utilize input and output warping~\cite{kumaraswamy1980generalized, box1964analysis} to map data distributions to ones that closely resemble Gaussian densities and rather adopt a multi-objective acquisition to determine Pareto-optimal solution between $\alpha_{\text{EI}}(\cdot)$, $\alpha_{\text{PI}}(\cdot)$, and $\alpha_{\text{UCB}}(\cdot)$, i.e., solving $\max_{\bm{\theta} \in \Theta} (\alpha_{\text{EI}}(\cdot), \alpha_{\text{PI}}(\cdot), - \alpha_{\text{UCB}}(\cdot))$; see~\cite{cowenempirical}.

\section{SYSTEM ARCHITECTURE}

\subsection{Hardware \& Software}\label{Sec:Hardware}
To demonstrate the effectiveness of general robotic systems in tackling dynamic tasks, we created an air hockey game between two Kuka LBR IIWA 14 manipulators. We mounted those arms at either end of an air hockey table and equipped each with a custom-designed end-effector of length 515 mm. The end-effector was composed of an aluminum rod, a 10 N gas spring, and a universal joint connected to a mallet. Our choice of a gas spring reduced contact forces to the table in case of vertical control error and as such increased safety by diminishing exerted pressures from the robotic arms. Additionally, the universal joint passively adapts row and pitch angles of the end-effector to ensure that the mallet's surface is parallel to the table. The cylindrical symmetry of the mallet warrants collisions which are invariant to yaw angles. Additionally, we positioned six Optitrack Flex 13 cameras with $(1920 \times 1680)$ pixel resolution and frame rates of 120 Hz above the air hockey table. We enabled effective object tracking by placing distinctive markers on both the puck and the table allowing us to achieve a 1mm tracking precision. 

On the software side, we built our framework using ROS  
in both real-world and simulation as illustrated in Fig. \ref{fig:framework}. In the real system, the puck and table poses were determined using the Optitrack system, while in simulation those poses were directly provided by Gazebo. Finally, we introduced a referee node that established the game's status and score. This node handles game logic by observing the state of the robot and the puck, performed checks of faulty states, and reset the puck's position in case of a scored goal or when the puck was unreachable.

\subsection{State Estimation} 
With our hardware setting presented, we now focus on the procedure by which we gather puck data (i.e., position and velocity) that we later use for both fine-tuning the simulator and for optimizing hitting movements (Section~\ref{Sec:Hit}). Given that the puck's position, $\bm{r}_t$, and orientation $\phi_t$ are attainable from the Optitrack system, we adopt an Extended Kalman filter to estimate linear and angular velocities $\dot{\bm{r}}_t$ and $\dot{\phi}_t$.

To estimate these quantities, we use an Extended Kalman Filter with the following transition model
\begin{align*}
    \vr_{t+1} = &  \vr_t +  \dot{\vr}_t \Delta T + \bm{\varepsilon}_\vr, \nonumber\\
    \dot{\vr}_{t+1} = &  \dot{\vr}_t - (d\dot{\vr}_t + \vc(\dot{\vr})) \Delta T + \bm{\varepsilon}_{\dot{\vr}}, \nonumber\\
    \phi_{t+1} = & \mathrm{wrap\_angle}(\phi_{t} + \dot{\phi_{t}} \Delta T + \varepsilon_{\phi}), \nonumber \\
    \dot{\phi}_{t+1} = & \dot{\phi}_{t} + \varepsilon_{\dot{\phi}},
\end{align*}
where $\varepsilon_k$ is the (Gaussian) noise specific to the state variable $k$,  $\vc(\cdot)$ is the friction term vector where we define every component $i$ as
\begin{align*}
    c_i(\dot{\vr}) = \begin{cases}
    c, & \lvert \dot{r}_i \rvert >0\\
    0 & \text{otherwise}.
\end{cases}
\end{align*}
As the Optitrack system can also track rigid bodies' orientation, the observed variables are $\vr$ and $\phi$.
After each prediction step, we apply the collision model from~\cite{Partridge2000}. Every time a collision happens, we don't update the covariance matrix, as it is hard to model the bounce uncertainty. 

We use an ellipsoidal gating procedure for the Optitrack measure. To compute the gate, we use the innovation covariance matrix and a gate probability of 90\% i.e., we reject the 10\% of the possible measurement which belongs to the queue of our estimated distribution.
As the Optitrack is very unlikely to produce outliers, we interpret the outliers as an unexpected collision. For this reason, instead of rejecting the measurement, we reset the track using the new information and reset the state covariance matrix to the initial one i.e., the identity matrix.

\begin{figure}[t]
    \centering
    \resizebox{.43\textwidth}{4.5cm}{
\begin{tikzpicture}[scale=0.13]
\tikzstyle{every node}+=[inner sep=0pt, font=\scriptsize]
\draw [black] (16.1,-13.1) circle (3);
\draw (16.1,-13.1) node {Init};
\draw [black] (49.4,-29.9) circle (3);
\draw (49.4,-29.9) node {Ready};
\draw [black] (65,-13.1) circle (3);
\draw (65,-13.1) node {Smash};
\draw [black] (16.1,-29.9) circle (3);
\draw (16.1,-29.9) node {Home};
\draw [black] (33.1,-42.5) circle (3);
\draw (33.1,-42.5) node {Cut};
\draw [black] (33.1,-13.1) circle (3);
\draw (33.1,-13.1) node {Prepare};
\draw [black] (65,-41.8) circle (3);
\draw (65,-41.8) node {Repel};
\draw [black] (16.1,-16.1) -- (16.1,-26.9);
\fill [black] (16.1,-26.9) -- (16.6,-26.1) -- (15.6,-26.1);
\draw (15.6,-21.5) node [left] {Start};
\draw [black] (14.777,-10.42) arc (234:-54:2.25);
\draw (16.1,-5.85) node [above] {!Start};
\fill [black] (17.42,-10.42) -- (18.3,-10.07) -- (17.49,-9.48);
\draw [black] (13.42,-31.223) arc (324:36:2.25);
\draw (8.85,-29.9) node [left] {!OnTable};
\fill [black] (13.42,-28.58) -- (13.07,-27.7) -- (12.48,-28.51);
\draw [black] (19.1,-29.9) -- (46.4,-29.9);
\fill [black] (46.4,-29.9) -- (45.6,-29.4) -- (45.6,-30.4);
\draw (32.75,-29.4) node [above] {OnTable};
\draw [black] (34.833,-40.052) arc (141.87424:113.53425:30.365);
\fill [black] (46.59,-30.96) -- (45.66,-30.82) -- (46.06,-31.74);
\draw (35.99,-34.28) node [above] {!Defense};
\draw [black] (47.83,-32.454) arc (-34.99486:-69.59665:25.208);
\fill [black] (35.97,-41.62) -- (36.89,-41.81) -- (36.54,-40.88);
\draw (43.4,-40.44) node [below] {Defense};
\draw [black] (63.705,-15.805) arc (-28.12944:-57.62837:33.778);
\fill [black] (52,-28.41) -- (52.95,-28.4) -- (52.41,-27.56);
\draw (59.2,-24.33) node [right] {Done};
\draw [black] (35.825,-14.351) arc (62.3239:25.94524:28.538);
\fill [black] (35.83,-14.35) -- (36.3,-15.17) -- (36.77,-14.28);
\draw (43.58,-18.28) node [right] {Stuck};
\draw [black] (46.639,-28.731) arc (-116.18815:-155.54271:26.556);
\fill [black] (46.64,-28.73) -- (46.14,-27.93) -- (45.7,-28.83);
\draw (38.77,-24.86) node [left] {Done};
\draw [black] (49.4,-45.3) -- (49.4,-32.9);
\draw (49.4,-45.8) node [below] {Pause};
\fill [black] (49.4,-32.9) -- (48.9,-33.7) -- (49.9,-33.7);
\draw [black] (6.7,-13.1) -- (13.1,-13.1);
\draw (6.2,-13.1) node [left] {Stop};
\fill [black] (13.1,-13.1) -- (12.3,-12.6) -- (12.3,-13.6);
\draw [black] (34.423,-45.18) arc (54:-234:2.25);
\draw (33.1,-49.75) node [below] {Defense};
\fill [black] (31.78,-45.18) -- (30.9,-45.53) -- (31.71,-46.12);
\draw [black] (31.777,-10.42) arc (234:-54:2.25);
\draw (33.1,-5.85) node [above] {!Done};
\fill [black] (34.42,-10.42) -- (35.3,-10.07) -- (34.49,-9.48);
\draw [black] (50.573,-27.14) arc (154.06648:120.17572:29.652);
\fill [black] (62.33,-14.47) -- (61.39,-14.44) -- (61.89,-15.31);
\draw (58.98,-15.47) node [left] {CanSmash};
\draw [black] (62.086,-41.098) arc (-107.64166:-147.0326:20.971);
\fill [black] (50.85,-32.52) -- (50.86,-33.47) -- (51.7,-32.92);
\draw (53.07,-38.28) node [below] {Done};
\draw [black] (52.178,-31.029) arc (65.18989:40.13585:31.889);
\fill [black] (63.18,-39.42) -- (63.04,-38.49) -- (62.28,-39.13);
\draw (62.29,-34.12) node [above] {CanRepel};
\draw [black] (63.677,-10.42) arc (234:-54:2.25);
\draw (65,-5.85) node [above] {!Done};
\fill [black] (66.32,-10.42) -- (67.2,-10.07) -- (66.39,-9.48);
\draw [black] (66.323,-44.48) arc (54:-234:2.25);
\draw (65,-49.05) node [below] {!Done};
\fill [black] (63.68,-44.48) -- (62.8,-44.83) -- (63.61,-45.42);
\draw [black] (52.58,-29.077) arc (144:-144:1);
\fill [black] (52.58,-30.22) -- (52.93,-31.1) -- (53.52,-30.29);
\end{tikzpicture}
}
    \caption{State machine of the high-level policy}
    \label{fig:state_machine}
\end{figure}
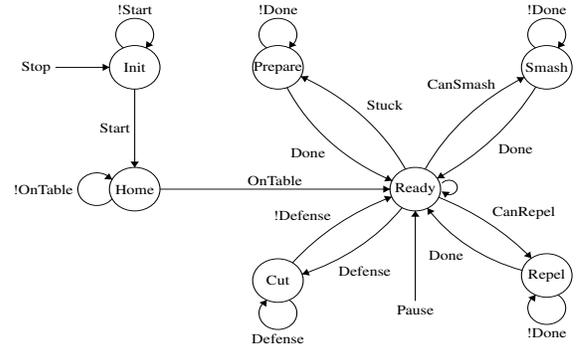

\begin{figure*}[t]
    \centering
        \begin{subfigure}[b]{.60\linewidth}
        \centering
        \includegraphics[width=\textwidth, trim=1.5cm 0.5cm 0.8cm 0.9cm, clip]{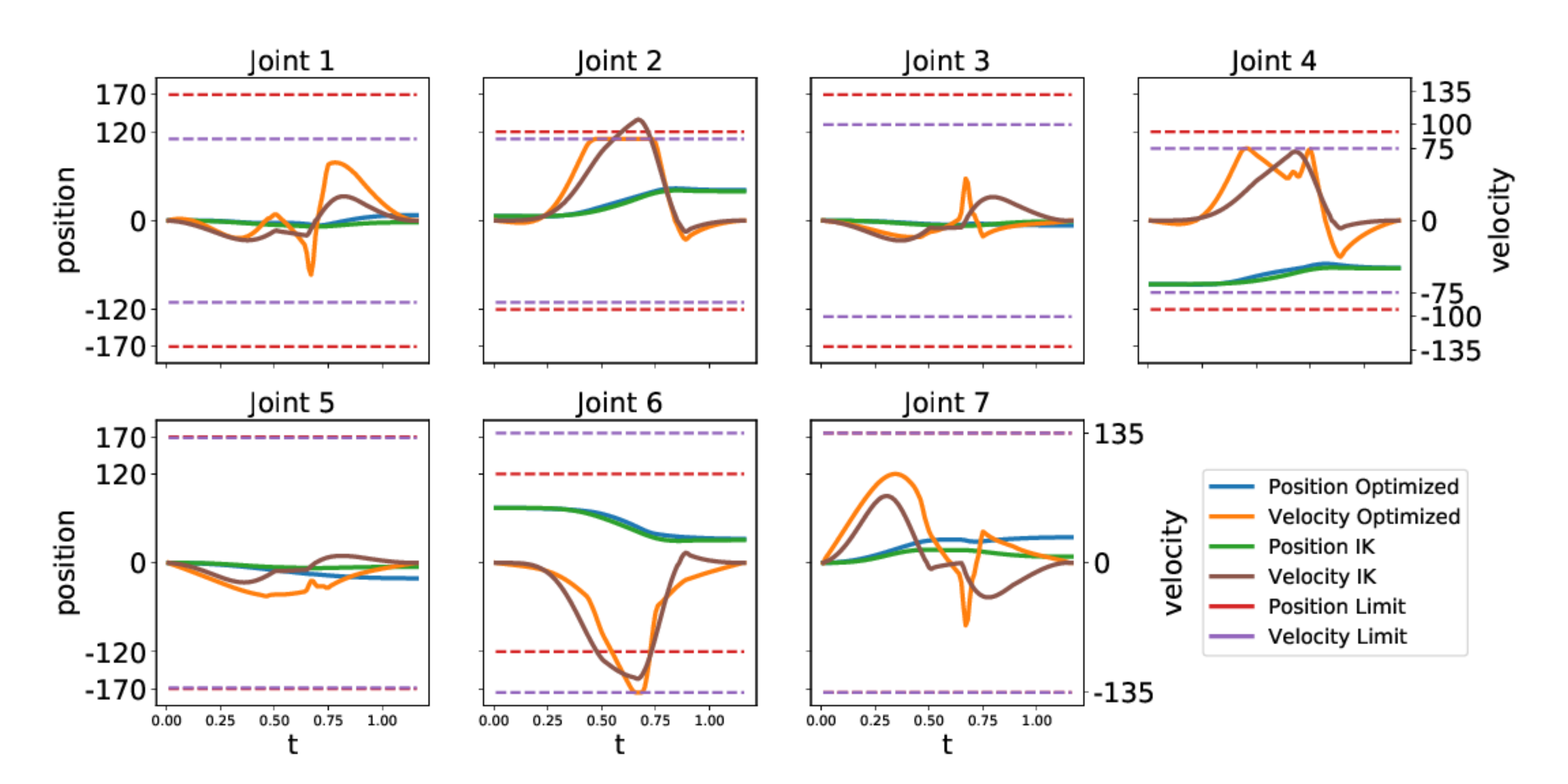}
        \caption{Optimized solution and inverse kinematic solution}
        \label{fig:opt_vs_IK}
    \end{subfigure}
    \hfill
    \begin{subfigure}[b]{.38\linewidth}
        \centering
        \includegraphics[width=\textwidth, trim=0cm 0.5cm 0.3cm 0.4cm, clip]{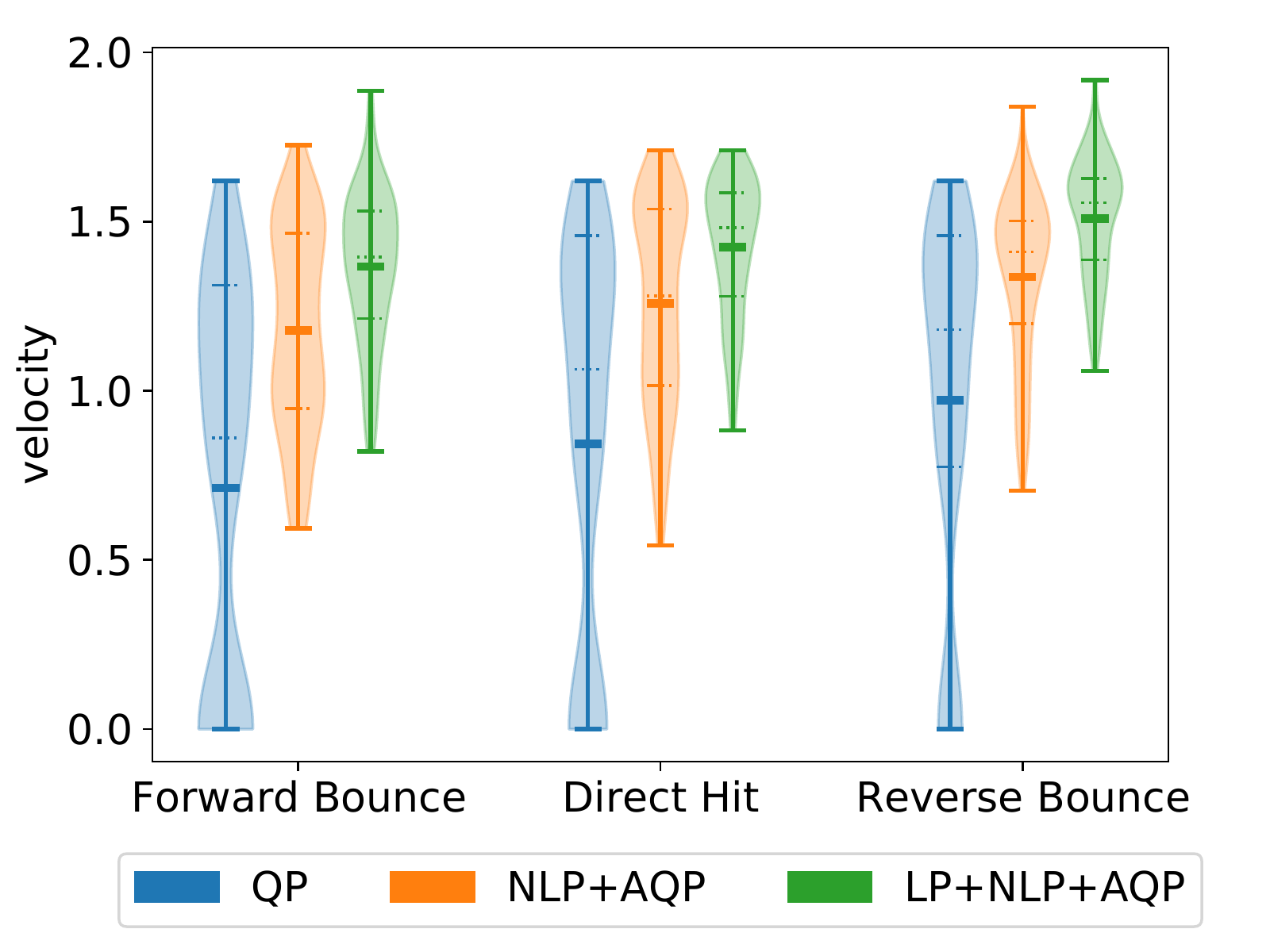}
        \caption{Hitting velocity for different algorithms}
        \label{fig:opt_res_with_failure}
    \end{subfigure}
    \caption{Hitting movement Optimization. (a) shows that the velocity limit gets violated on joint 2 for planning and inverse kinematic approach. Our approach apportions the velocity of joint 2 into other joints. (b) shows that \glsentryshort{nl} signific antly improves the hitting performance and eliminates the failures. \glsentryshort{lp} tries to reach the theoretical maximum of the configuration.}
    \label{fig:hitting}
\end{figure*}

\subsection{Agent Strategy}
Following previous works~\cite{shimada2017two}, we use a two-layer hierarchical architecture to control the robot. At a high level, we leverage a classical finite state machine to implement the tactics. 
The high-level strategies that we propose are very similar to the one presented in the previous literature~\cite{shimada2017two, AlAttar2019}. We show the full state machine in Fig.~\ref{fig:state_machine}. The \textit{Init} state is a safe state where the robot positions the mallet at a safe height from the table. When we stop the game, the robot reverts to this safe state, where no action can occur. Once we start the game, the robot enters the \textit{Home} state, where it moves the mallet on the table. Once both robots have positioned their mallets on the table, The system enters the \textit{Ready} state, which keeps the robot on the base position until an event occurs. The \textit{Ready} state is active while the game is paused. 

All the other possible states are reachable from the \textit{Ready} state.
The \textit{Smash} state can be activated when the puck velocity is not too high. It hits the puck strongly and aims at the goal. 
The \textit{Repel} state quickly responds to a fast incoming puck that is not risky, by hitting back the puck in the incoming direction. The \textit{Prepare} state is used when the puck is stationary and close to the table's borders.  It moves the puck in a better position to be able to hit it strongly. Finally, the \textit{Cut} state is responsible for defending the goal. It tries to move the incoming puck at the sides to gain control of it. 

\section{EXPERIMENTAL EVALUATION}
In this section, we present the experimental results for the hitting movement and the system identification procedure. In the attached video, we demonstrate the result of the Kalman filter, the system identification, the state machine, two robots playing the game autonomously in simulation, and the hitting performance on the real robot. Demonstrations can be found in the 
\href{https://sites.google.com/view/robot-air-hockey}{video}.
\subsection{Hitting Performance}
\begin{table}[b]
    \footnotesize
    \centering
    \begin{tabular}{|c|c|c|c|c|c|c|c|}
        \hline
         Method & Mean & Min & Max & Median \\
         \hline
         \glsentryshort{qp} & 263.58 & 30.65 & 531.73 & 243.96\\
         \hline
         \glsentryshort{nl} + \glsentryshort{aqp} & 132.88 & 23.03 & 284.54 & 126.07 \\
         \hline
         \glsentryshort{lp} + \glsentryshort{nl} + \glsentryshort{aqp} & 234.26 & 43.26 & 498.84 & 222.01\\
         \hline
    \end{tabular}
    \caption{Optimization time (\textit{ms}) for different methods}
    \label{tab:optimization_time}
\end{table}

In Fig.~\ref{fig:opt_vs_IK} we compare the trajectories obtained using only inverse kinematics on the planned path with the ones optimized by \gls{lp} + \gls{nl} + \gls{aqp}. The proposed optimization makes better use of the velocity range of each joint while obtaining similar joints paths. This technique exploits the null space manifold such that we avoid exceeding the velocity bounds.
In the shown example, instead of overshooting the velocity limit on joint 2, the optimized trajectory apportions the velocity to other joints. We can also observe that multiple joints are reaching the velocity limit at the same time, which means robot capability gets exploited.

The violin plot in Fig.~\ref{fig:opt_res_with_failure} shows the performance of each component in the proposed optimization. We selected three hitting types i.e., hitting directly towards the goal, make a bounce towards the opposite side (reverse bounce), or the same side of the puck (forward bounce). For each hitting type, we evaluate the movement on a grid of 180 puck hitting positions. The trajectories are planned with the time step size of 0.01s. In Table~\ref{tab:optimization_time} is reported the elapsed time statistics for every algorithm. For every hitting point, we perform up to 10 trials of hitting optimization. For \gls{qp}, the initial attempting velocity is set to be $2m/s$, other methods use the velocity solved from Section~\ref{sec:VelOpt}. If the optimization succeeds, we stop the optimization and obtain the calculation time. If the optimization fails, the velocities are scaled down by a factor of 0.9 and retry. If all trials fail, we record the hitting velocity as zero and the optimization time at termination. 

From the violin plots, it is clear that the initial nonlinear optimization (NL) \eqref{eq:nl_opt} together with the anchored quadratic programming (AQP) \eqref{eq:anchor_opt} drastically improves the chances of successfully compute a hitting trajectory compared to the pure quadratic programming (QP) approach using \eqref{eq:null_space_opt}. We record no hitting failure in the tested position with NL. The linear programming step (LP) shown in \eqref{eq:linear_prog} to compute the maximum velocity allows us to find fast-hitting solutions at the cost of double computation time. This higher computation time is mostly due to the failure of optimization, because there is no feasible solution to reach the desired velocity following the Cartesian path. By employing a reasonable prediction time, e.g., 1.5 seconds, it is possible to use any of the proposed algorithms to hit the puck, as the optimization, except for the most extreme positions, is sufficiently fast to allow the hitting movement of the robot. This issue is not crucial in practical cases: when the optimization takes longer it is not possible to obtain a high hitting velocity therefore it may not be worth trying the hitting movement in the first place. This fact allows us to play the game reactively with the robot, as proved in simulation.

\begin{figure*}[t]
    \centering
        \begin{subfigure}[t]{0.24\textwidth}
        \includegraphics[width=\textwidth, 
        trim=0.2cm 0cm 0.5cm 0cm, clip] {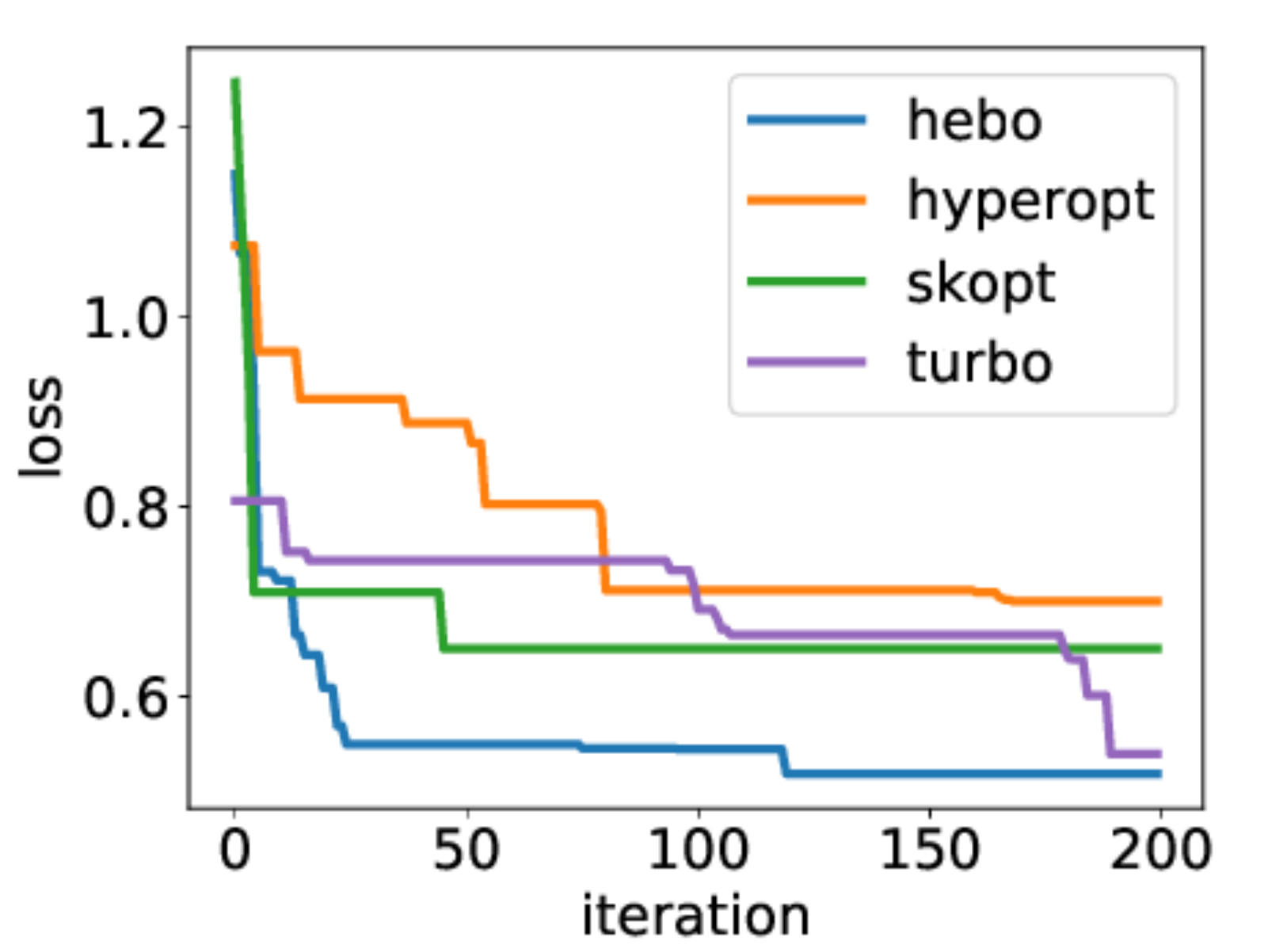}
        \caption{Learning curve}
        \label{fig:bo_learning_curve}
    \end{subfigure}
    \hfill
    \begin{subfigure}[t]{0.24\textwidth}
        \includegraphics[width=\textwidth, 
        trim=0.2cm 0cm 0.5cm 0cm, clip] {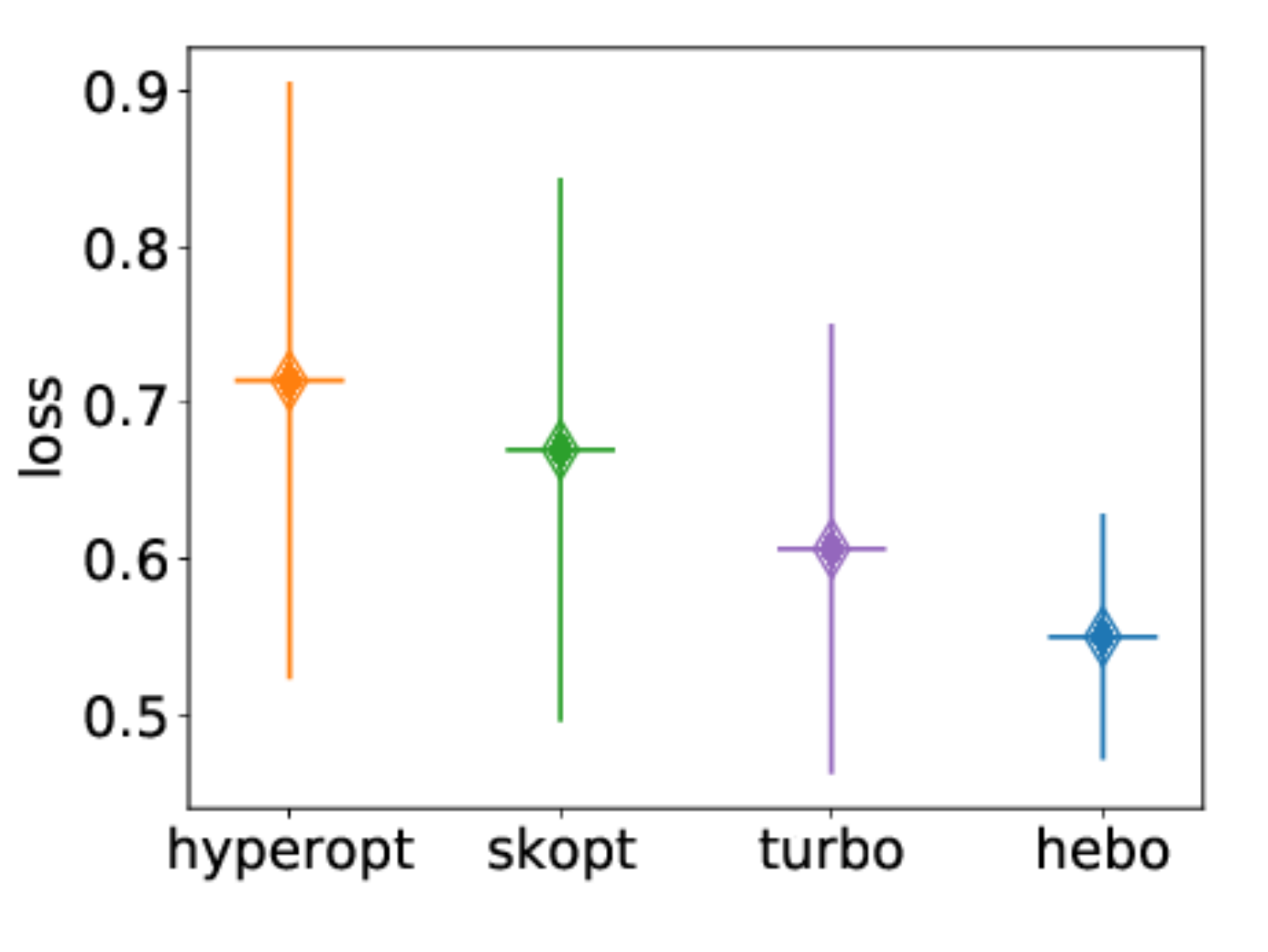}
        \subcaption{Best results over 3 runs, with 95\% confidence intervals}
        \label{fig:bo_best}
    \end{subfigure}
    \hfill
    \begin{subfigure}[t]{0.24\textwidth}
        \includegraphics[width=\textwidth, 
        trim=0cm 0cm 0 0.6cm, clip]
        {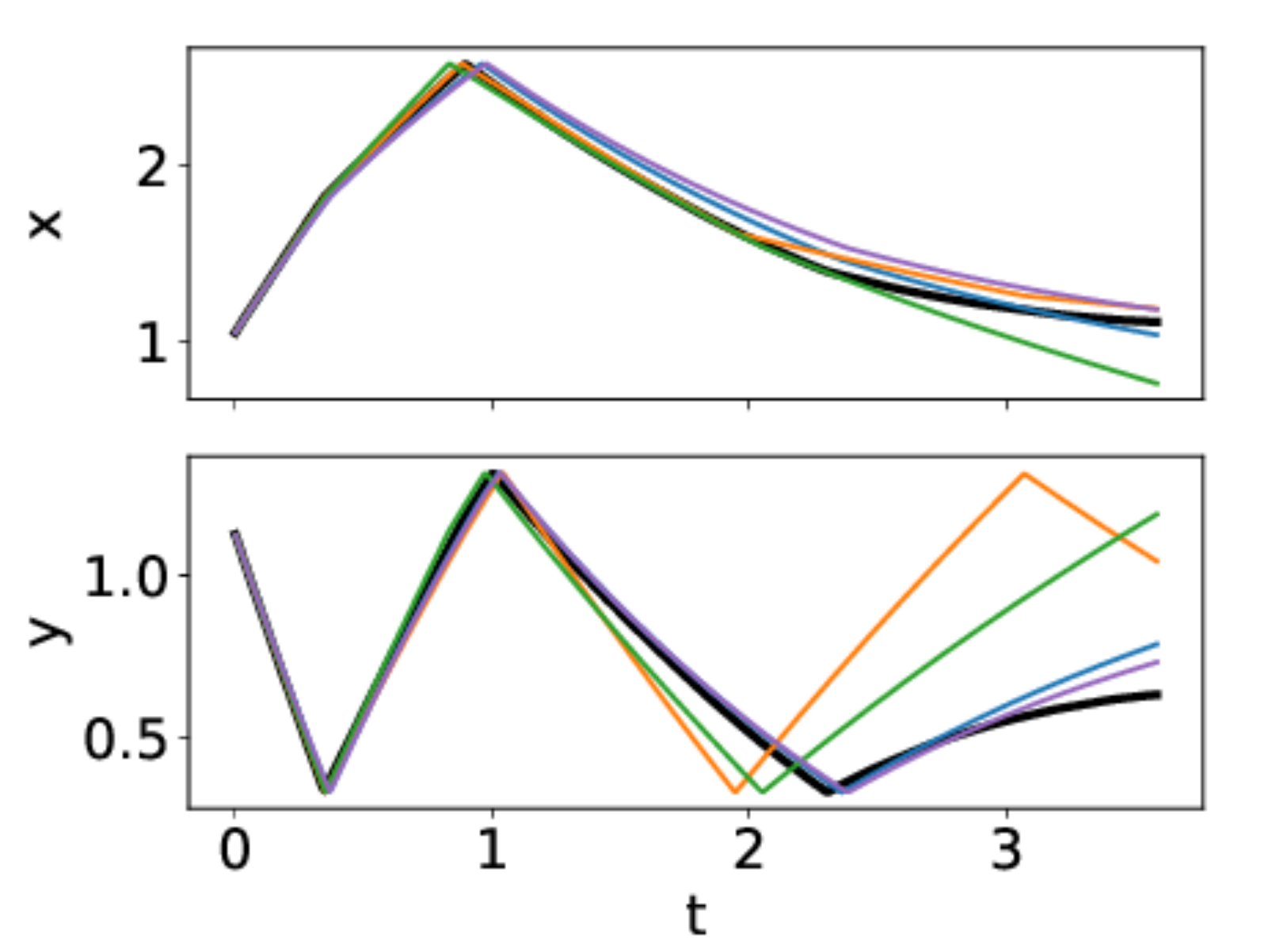}
        \subcaption{Example trajectory in time}
        \label{fig:bo_trajectory_t}
    \end{subfigure}
    \hfill
    \begin{subfigure}[t]{0.24\textwidth}
        \centering
        \includegraphics[width=\textwidth, 
        trim=0cm 0 0.5cm 0.5cm, 
        clip]{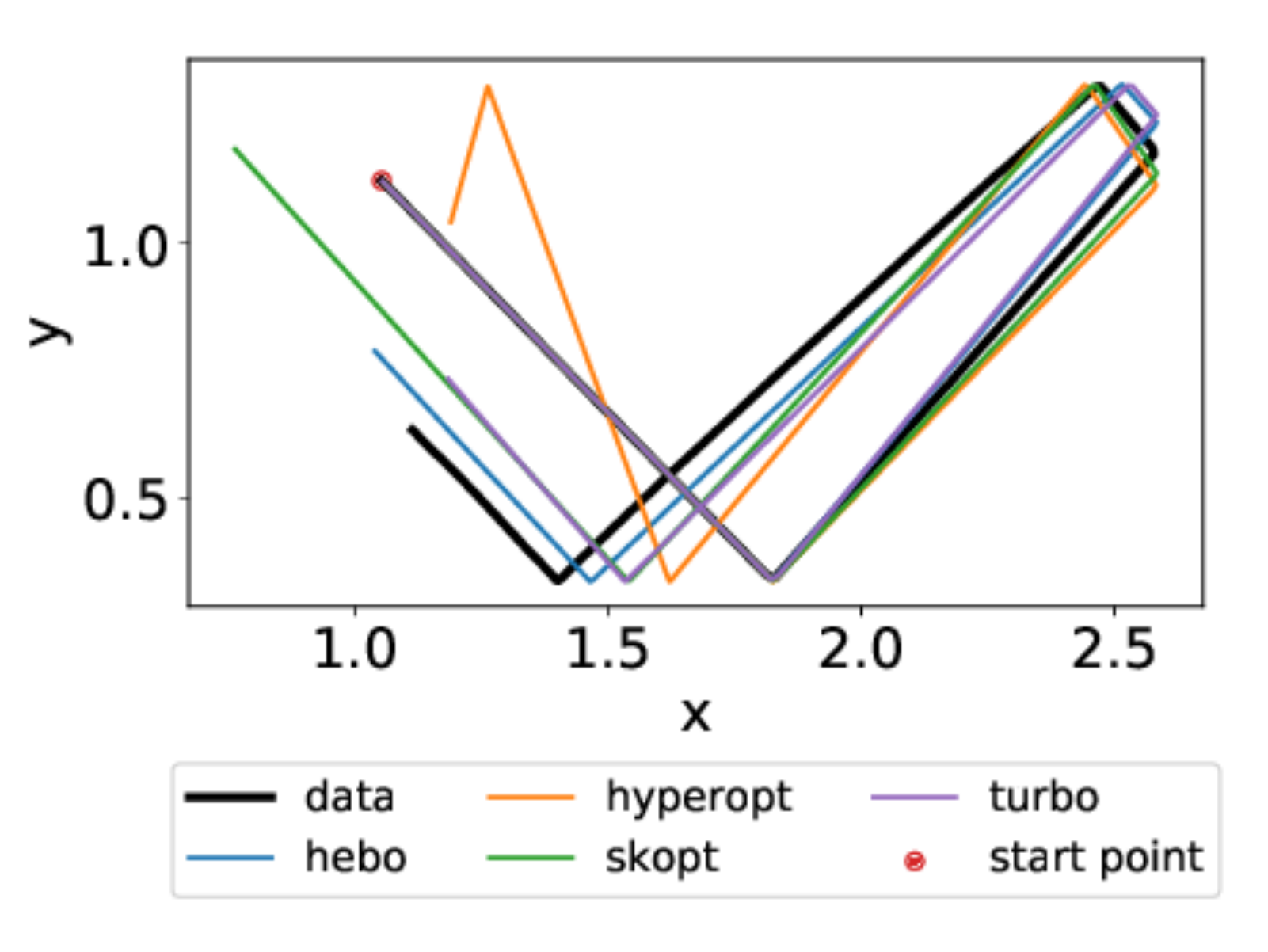}
        \subcaption{Example trajectory in x-y plane}
        \label{fig:bo_trajectory_xy}
    \end{subfigure}
    \caption{Results for Bayesian optimization.}
    \label{fig:bo_results}
\end{figure*}
\subsection{System Identification}
We optimize the parameters for 200 iterations. At each iteration, we sample 5 candidate parameter vectors from the acquisition function and we evaluate the loss over the whole dataset. In our experiments, \gls{hebo} significantly outperformed all other \gls{bo} algorithms, leading to trajectories that are closer to the ones in the training dataset, as is shown in Fig.~\ref{fig:bo_trajectory_t} and~\ref{fig:bo_trajectory_xy}.
From the learning curve in Fig.~\ref{fig:bo_learning_curve}, it is clear that not only \gls{hebo} finds a better solution faster than the other algorithms, but also the samples it selects are on average better than the ones chosen by competing algorithms. We also observe superior performances when averaging among different independent runs, as presented in Fig~\ref{fig:bo_best}.
While the results are sufficiently good to be used in a simulation, there is still some discrepancy between the real trajectory and the simulated one. This discrepancy is unavoidable since it's impossible to have a perfect model of the air hockey table. Indeed, irregularities of the plane and air flow play a major role in the system, particularly at slow velocities, as Fig.~\ref{fig:bo_trajectory_t} shows.

\section{CONCLUSION}
In this paper, we studied and presented a robotic system that is able to play air hockey, a simple, but challenging dynamic task. We proved that commercially available general-purpose industrial manipulators, such as the Kuka Iiwa LBR 14, can perform strong hits. We proposed an optimization method that exploits the redundancy of the manipulator coping with joint velocity limits. In the simulated environment, we showed two robots playing against each other successfully, while our optimization algorithm runs in real-time on a desktop machine. We presented a simple framework, based on \gls{bo}, to tune the parameters of the environment's dynamic model. This framework only requires black-box access to a simulator: it doesn't require any further knowledge of the system.

This work opens many interesting research lines. An interesting research direction is to perform an in-depth study of \gls{bo} techniques applied to non-differentiable dynamics, such as the collision, and compare with standard techniques. Another research direction would be to improve the optimization by considering dynamics properties, and not only kinematics.
Finally,  we need to test the whole game on the real system to demonstrate the capabilities of existing systems and our approach.





\section*{ACKNOWLEDGMENT}
This project was supported by the CSTT fund from Huawei Tech R\&D (UK). The support provided by China Scholarship Council (No. 201908080039) is acknowledged.


\bibliographystyle{ieeetr}
\bibliography{bibliography}

\begin{thebibliography}{10}

\bibitem{Partridge2000}
C.~B. Partridge and M.~W. Spong, ``Control of planar rigid body sliding with
  impacts and friction,'' {\em International Journal of Robotics Research
  (IJRR)}, vol.~19, no.~4, pp.~336--348, 2000.

\bibitem{Namiki2013}
A.~Namiki, S.~Matsushita, T.~Ozeki, and K.~Nonami, ``Hierarchical processing
  architecture for an air-hockey robot system,'' in {\em IEEE International
  Conference on Robotics and Automation (ICRA)}, 2013.

\bibitem{shimada2017two}
H.~Shimada, Y.~Kutsuna, S.~Kudoh, and T.~Suehiro, ``A two-layer tactical system
  for an air-hockey-playing robot,'' in {\em IEEE/RSJ International Conference
  on Intelligent Robots and Systems (IROS)}, 2017.

\bibitem{AlAttar2019}
A.~AlAttar, L.~Rouillard, and P.~Kormushev, ``Autonomous air-hockey playing
  cobot using optimal control and vision-based bayesian tracking,'' in {\em
  International Conference Towards Autonomous Robotic Systems (TAROS)}, 2019.

\bibitem{bishop1999vision}
B.~E. Bishop and M.~W. Spong, ``Vision based control of an air hockey playing
  robot,'' {\em IEEE Control Systems Magazine}, vol.~19, no.~3, 1999.

\bibitem{Alizadeh2013}
H.~Alizadeh, H.~Moradi, and M.~N. Ahmadabadi, ``Automatic calibration of an air
  hockey robot,'' in {\em International Conference on Robotics and Mechatronics
  (ICRoM)}, pp.~107--112, IEEE, 2013.

\bibitem{Igeta2017}
K.~Igeta and A.~Namiki, ``Algorithm for optimizing attack motions for
  air-hockey robot by two-step look ahead prediction,'' in {\em IEEE/SICE
  International Symposium on System Integration}, pp.~465--470, 2017.

\bibitem{Taitler2017}
A.~{Taitler} and N.~{Shimkin}, ``Learning control for air hockey striking using
  deep reinforcement learning,'' in {\em International Conference on Control,
  Artificial Intelligence, Robotics Optimization}, 2017.

\bibitem{mnih2015human}
V.~Mnih, K.~Kavukcuoglu, D.~Silver, A.~A. Rusu, J.~Veness, M.~G. Bellemare,
  A.~Graves, M.~Riedmiller, A.~K. Fidjeland, G.~Ostrovski, {\em et~al.},
  ``Human-level control through deep reinforcement learning,'' {\em nature},
  vol.~518, no.~7540, pp.~529--533, 2015.

\bibitem{Xie2020}
A.~Xie, D.~P. Losey, R.~Tolsma, C.~Finn, and D.~Sadigh, ``Learning latent
  representations to influence multi-agent interaction,'' in {\em Conference on
  Robot Learning (CoRL)}, 2020.

\bibitem{Carl}
C.~E. Rasmussen and C.~K.~I. Williams, {\em Gaussian Processes for Machine
  Learning (Adaptive Computation and Machine Learning)}.
\newblock The MIT Press, 2005.

\bibitem{movckus1975bayesian}
J.~Mo{\v{c}}kus, ``On bayesian methods for seeking the extremum,'' in {\em
  Optimization techniques IFIP technical conference}, Springer, 1975.

\bibitem{kushner1964new}
H.~J. Kushner, ``A new method of locating the maximum point of an arbitrary
  multipeak curve in the presence of noise,'' {\em Journal of Basic
  Engineering}, vol.~86, no.~1, pp.~97--106, 1964.

\bibitem{srinivas2009gaussian}
N.~Srinivas, A.~Krause, S.~M. Kakade, and M.~Seeger, ``Gaussian process
  optimization in the bandit setting: No regret and experimental design,'' {\em
  arXiv preprint arXiv:0912.3995}, 2009.

\bibitem{biagiotti2008trajectory}
L.~Biagiotti and C.~Melchiorri, {\em Trajectory planning for automatic machines
  and robots}.
\newblock Springer Science \& Business Media, 2008.

\bibitem{vahrenkamp2012manipulability}
N.~Vahrenkamp, T.~Asfour, G.~Metta, G.~Sandini, and R.~Dillmann,
  ``Manipulability analysis,'' in {\em IEEE-RAS International Conference on
  Humanoid Robots (HUMANOIDS)}, pp.~568--573, IEEE, 2012.

\bibitem{cowenempirical}
A.~I. Cowen-Rivers, W.~Lyu, R.~Tutunov, Z.~Wang, A.~Grosnit, R.~R. Griffiths,
  H.~Jianye, J.~Wang, and H.~B. Ammar, ``An empirical study of assumptions in
  bayesian optimisation.'' unpublished.

\bibitem{kumaraswamy1980generalized}
P.~Kumaraswamy, ``A generalized probability density function for double-bounded
  random processes,'' {\em Journal of hydrology}, 1980.

\bibitem{box1964analysis}
G.~E. Box and D.~R. Cox, ``An analysis of transformations,'' {\em Journal of
  the Royal Statistical Society: Series B (Methodological)}, 1964.

\end{thebibliography}

\end{document}